\documentclass[letterpaper, 10 pt, conference]{ieeeconf}
\IEEEoverridecommandlockouts
\usepackage{microtype}
\usepackage{graphicx}
\usepackage{booktabs}
\usepackage{verbatim}

\usepackage{amsmath}
\usepackage{amssymb}
\usepackage{mathtools}
\usepackage{amsfonts}
\usepackage{stackengine}
\usepackage{xspace}

\usepackage{cite}
\usepackage{multicol}
\usepackage{subcaption}
\usepackage[dvipsnames]{xcolor}
\usepackage{multirow}
\usepackage{bm}
\usepackage{float}
\usepackage{etoolbox}
\usepackage{wrapfig}
\usepackage{cancel}
\usepackage{times}
\usepackage{etoolbox}

\makeatletter
\let\NAT@parse\undefined
\makeatother
\usepackage{hyperref} 

\usepackage[capitalize,noabbrev]{cleveref}

\DeclareMathOperator*{\argmax}{arg\,max}
\DeclareMathOperator*{\argmin}{arg\,min}

\newcommand{\Expect}[2]{\mathbb{E}_{#1}\Big[#2\Big]}
\newcommand{\Expectt}[2]{\mathbb{E}_{#1}\Bigg[#2\Bigg]}

\newcommand{\Par}[1]{\Big( #1 \Big)}
\newcommand{\real}{\mathbb{R}}

\newcommand{\relu}{\text{ReLU}}

\newcommand{\hrulethick}{\specialrule{0.1em}{0em}{0em}}

\newcommand{\Ncal}{\mathcal{N}}
\newcommand{\Mcal}{\mathcal{M}}

\newcommand{\Xcal}{\mathcal{X}}
\newcommand{\Ucal}{\mathcal{U}}
\newcommand{\Hcal}{\mathcal{H}}

\newcommand{\mutilde}{\tilde{\mu}}

\newcommand{\xbold}{\bm{x}}
\newcommand{\ubold}{\bm{u}}
\newcommand{\pbold}{\bm{p}}
\newcommand{\vbold}{\bm{v}}
\newcommand{\gbold}{\bm{g}}
\newcommand{\ebold}{\bm{e}}
\newcommand{\Rbold}{\bm{R}}

\newcommand{\qbold}{\bm{q}}

\newcommand{\xhatbold}{\bm{\hat{x}}}
\newcommand{\uhatbold}{\bm{\hat{u}}}

\newcommand{\uhat}{\hat{u}}

\newcommand{\Jhat}{\hat{J}}

\newcommand{\pibold}{\bm{\pi}}
\newcommand{\mubold}{\bm{\mu}}

\newcommand{\omegabold}{\bm{\omega}}
\newcommand{\sigmabold}{\bm{\sigma}}
\newcommand{\Sigmabold}{\bm{\Sigma}}
\newcommand{\thetabold}{\bm{\theta}}

\newcommand{\dmpo}{\texttt{DMPO}\xspace}
\newcommand{\mppi}{\texttt{MPPI}\xspace}
\newcommand{\ete}{\texttt{E2E}\xspace}

\definecolor{myblue}{RGB}{119, 177, 233}
\definecolor{mypurple}{RGB}{212, 152, 229}
\definecolor{mygreen}{RGB}{184, 233, 134}
\definecolor{myyellow}{RGB}{233, 213, 134}

\title{\LARGE \bf
Deep Model Predictive Optimization
}

\author{Jacob Sacks$^{*}$, Rwik Rana$^{*}$, Kevin Huang$^{*}$, Alex Spitzer$^{*}$, Guanya Shi$^{\dagger}$, and Byron Boots$^{*}$
\thanks{$^{*}$University of Washington, Seattle WA 98105, USA. $^\dagger$Robotics Institute, Carnegie Mellon University, Pittsburgh PA 15213, USA.
Email: {\tt\small jsacks6@cs.washington.edu}}%
}

\begin{document}
\maketitle

\begin{abstract}
A major challenge in robotics is to design robust policies which enable complex and agile behaviors in the real world.
On one end of the spectrum, we have model-free reinforcement learning (MFRL), which is incredibly flexible and general but often results in brittle policies.
In contrast, model predictive control (MPC) continually re-plans at each time step to remain robust to perturbations and model inaccuracies.
However, despite its real-world successes, MPC often under-performs the optimal strategy.
This is due to model quality, myopic behavior from short planning horizons, and approximations due to computational constraints.
And even with a perfect model and enough compute, MPC can get stuck in bad local optima, depending heavily on the quality of the optimization algorithm.
To this end, we propose Deep Model Predictive Optimization (\dmpo), which learns the inner-loop of an MPC optimization algorithm directly via experience, specifically tailored to the needs of the control problem.
We evaluate \dmpo on a real quadrotor agile trajectory tracking task, on which it improves performance over a baseline MPC algorithm for a given computational budget.
It can outperform the best MPC algorithm by up to 27\% with fewer samples and an end-to-end policy trained with MFRL by 19\%.
Moreover, because \dmpo requires fewer samples, it can also achieve these benefits with $4.3 \times$ less memory.
When we subject the quadrotor to turbulent wind fields with an attached drag plate, \dmpo can adapt zero-shot while still outperforming all baselines.
Additional results can be found at \href{https://tinyurl.com/mr2ywmnw}{https://tinyurl.com/mr2ywmnw}.
\end{abstract}

\section{Introduction}

For robots to perform complex and agile behaviors in the real world, it is crucial to design control policies that remain robust while pushing the limits of the system.
Model-free reinforcement learning (MFRL) is a general approach that makes minimal assumptions on the problem and has been successfully deployed in the real world \cite{openai2018learning, huang2023datt, kaufmann2023champion}.
However, policies trained with these methods are often brittle and do not generalize to out-of-distribution disturbances.
For instance, consider an uninhabited aerial vehicle (UAV) following an aggressive trajectory in uncertain environments~\cite{o2022neural,shi2019neural}.
If the UAV encounters unknown wind gusts that were not experienced in training, the policy will likely not be able to account for the change in dynamics and lead to a crash.
Furthermore, due to the sample inefficiency of MFRL methods, they often train policies in simulation.
This can create a sim-to-real gap due to the mismatch between the simulator and the true system.
Even if the policies succeed in simulation, they often fail in the real world.
We can partially remedy this issue using domain randomization (DR) \cite{tobin2017domain, peng2018sim, chen2021understanding}.
However, its efficacy is dependant on the parameter distributions we select and the nature of the problem.

Alternatively, model predictive control (MPC) is a powerful framework which leverages a model to iteratively re-plan a finite-horizon control sequence at each time step~\cite{morari1999model,williams2017information,yu2020power}.
The first control from this plan is applied to the system and the process repeats.
Although the planned sequence is often open-loop, because we are updating it using the current state, MPC effectively yields a state-feedback policy.
By re-solving this optimization problem online, we can improve the robustness of MPC to perturbations and model inaccuracies.
However, this also leads to increased computational demands compared to MFRL.
In practice, we approximate the solution at each time step to run in real time.
This involves warm-starting with the solution from the previous time step, which works well when the two problems are similar \cite{wagener2019online}.
But if our system encounters a large perturbation, this warm-starting procedure can bias us towards a poor solution \cite{erez2012infinite}.
The performance of MPC also depends on the model quality and prediction horizon length~\cite{yu2020power, jiang2015dependence, tamar2017learning, li2022robustness}.
And even with a perfect model and enough computation, the optimization algorithm may get stuck in bad local optima \cite{jain2021optimal}.
Altogether, these issues often lead to MPC under-performing the optimal policy.

To improve the performance of MPC while retaining its robustness, we propose Deep Model Predictive Optimization (\dmpo), which \emph{learns} an optimizer and warm-starting procedure directly via experience.
That is, \dmpo learns how to perform model-based planning more effectively while considering the computational demands for real-time deployment.
Our key contributions are:
\begin{enumerate}
    \item We develop \dmpo, a general approach for learning the inner-loop of the MPC optimizer and warm-starting procedure via reinforcement learning by viewing MPC as a structured recurrent policy class;
    \item On a real quadrotor platform (Crazyflie 2.1 with upgraded motors) tracking infeasible zig-zag trajectories, we show that \dmpo can outperform an end-to-end (\ete) policy trained with MFRL by 19\%;
    \item Tracking zig-zag trajectories with alternating $180^o$ flips in the desired yaw, \dmpo can improve error over a baseline MPC algorithm by up to 27\% with $16\times$ fewer samples, saving $4.3 \times$ memory requirements;
    \item By exposing the quadrotor to turbulent wind fields with an attached cardboard drag plate, we show that \dmpo can adapt zero-shot, matching the performance of the MPC baseline and outperforming \ete by 14\%.
\end{enumerate}


\section{Preliminaries}
\subsection{Reinforcement Learning}
We consider controlling a discrete-time stochastic dynamical system as an infinite-horizon discounted Markov decision process (MDP) defined by the tuple $\Mcal = (\Xcal, \Ucal, P, r, \rho_0, \gamma)$, where $\Xcal$ is the state space, $\Ucal$ is the control space, $x_{t+1} \sim P(\cdot | x_t, u_t)$ is the transition dynamics, $r(x, u)$ is the reward function, $\rho_0(x_0)$ is the initial state distribution, and $\gamma \in (0, 1)$ is the discount factor.
Given a closed-loop policy, $u \sim \pi(\cdot | x)$, its value function is defined as
\begin{equation}
V^{\pi}(x) = \Expectt{P, \pi}{\sum_{t=0}^{\infty} \gamma^t r(x_{t}, u_{t}) \Big | x_0 = x}.
\end{equation}
%
The goal of reinforcement learning (RL) is to find a policy which maximizes the expected discounted reward, which is equivalent to maximizing the value function:
\begin{equation}
\pi^* = \argmax_{\pi} \Expect{x_0 \sim \rho_0}{V^{\pi}(x_0)}.
\end{equation}
A common approach is to directly find $\pi$ with policy gradient methods, which perform gradient descent with a zeroth-order approximation of the gradient using a finite number of samples.
State-of-the-art approaches include actor-critic algorithms \cite{schulman2017proximal, schulman2015trust}, which in addition to learning $\pi$ (actor), learn an estimate of $V_{\pi}$ (critic) and use it as a baseline to reduce the gradient estimator variance.
If we wish to train our policy over a range of tasks, we can additionally condition both $\pi$ and $V_{\pi}$ on task parameters, such as a goal state.

\subsection{Sampling-based Model Predictive Control}
Rather than find a single, globally optimal policy, MPC re-optimizes a local policy at each time step. 
It accomplishes this by predicting the system's behavior over a finite horizon $H$ using an approximate model $\hat P$.
In sampling-based MPC, this local policy is often a distribution over open-loop control sequences, $\uhatbold_t \sim \pibold_{\thetabold_t}(\cdot)$, where $\uhatbold_t \triangleq (\hat u_t, \hat u_{t+1}, \dots, \hat u_{t+H-1})$ and $\thetabold_t \in \Theta$ are some set of feasible parameters.
At each time step, we solve
$\thetabold_t \leftarrow \argmin_{\thetabold \in \Theta} \Jhat(\thetabold; x_t)$,
where $\Jhat(\thetabold; x_t)$ is a statistic defined on $C(\xhatbold_t, \uhatbold_t)$, the total cost of our predicted trajectory over the finite horizon $H$.
After finding the solution, we sample a control sequence from our new policy and apply the first control, $u_t = \uhat_t$.
Despite the plan being open-loop, the MPC procedure can be thought of as outlining a state-feedback policy.
This is because we are updating the open-loop sequence using information about the current state.

A popular sampling-based approach to MPC is Model Predictive Path Integral (\mppi) control \cite{williams2016aggressive, williams2017information}, which assumes that our policy is a factorized Gaussian of the form
\begin{equation}
\pibold_{\thetabold}(\uhatbold) = \prod_{h=0}^{H-1} \pi_{\theta_h}(\uhat_{t+h})
= \prod_{h=0}^{H-1}  \Ncal(\uhat_{t+h}; \mu_{t+h}, \Sigma_{t+h}).
\label{eq:factorized_gaussian}
\end{equation}
MPPI also assumes that we optimize the exponential utility of our cost function, defined as:
\begin{equation}
\Jhat(\thetabold; x_t) = - \log \Expect{\pi_{\thetabold}, \hat P}{\exp \Par{-\frac{1}{\beta} C(\xhatbold_t, \uhatbold_t)}},
\end{equation}
where $\beta>0$ is a scaling parameter, known as the temperature.
We can approximate the gradient of this objective with samples and compute an update via dynamic mirror descent (DMD) \cite{wagener2019online} to arrive at the \mppi update rule:
\begin{equation}
\mu_{t+h} = (1-\gamma_t^{\mu}) \tilde{\mu}_{t+h} + \gamma_t^{\mu} \sum_{i=1}^N w_i \uhat_{t+h}^{(i)},
\label{eq:mppi_mean_update}
\end{equation}
\begin{equation}
\Sigma_{t+h} = (1-\gamma_t^\sigma) \tilde{\Sigma}_{t+h} + \gamma_t^\sigma \sum_{i=1}^N w_i m_{t+h}^{(i)}m_{t+h}^{{(i)}^T},
\label{eq:mppi_variance_update}
\end{equation}
where $\mutilde_{t+h}$ and $\tilde \Sigma_{t+h}$ are the current mean and covariance matrix for each time step, $m_{t+h} = u_{t+h} - \mu_{t+h}$, $\gamma_t^\mu$ and $\gamma_t^\sigma$ are the corresponding step sizes, and the weights are:
\begin{equation}
w_i = \frac{
e^{-\frac{1}{\beta} C(\xhatbold_t^{(i)}, \uhatbold_t^{(i)})}
}{
\sum_{j=1}^N e^{-\frac{1}{\beta} C(\xhatbold_t^{(j)}, \uhatbold_t^{(j)})}
}.
\label{eq:mppi_weights}
\end{equation}
Despite its robustness, MPC often requires a large number of sampled trajectories or multiple update iterations, which is infeasible due to real-time constraints.
To improve convergence, we can initialize the current parameters of the optimization problem as a function of the previous approximate solution, $\tilde \thetabold_{t+1} = \Phi(\thetabold_t)$, where $\Phi$ is called the shift model.
A common choice is to shift the parameter sequence forward by one time step.
For instance, due to \Cref{eq:factorized_gaussian}, $\thetabold_{t} = (\theta_{t}, \theta_{t+1}, \dots, \theta_{t+H-1})$ and $\tilde{\thetabold}_{t+1} = (\theta_{t+1}, \theta_{t+2}, \dots, \theta_{t+H-1}, \bar{\theta})$, where $\bar{\theta}$ is a hyperparameter.

\begin{figure*}[t]
\centering
\vspace{2ex}
\includegraphics[width=0.9\textwidth]{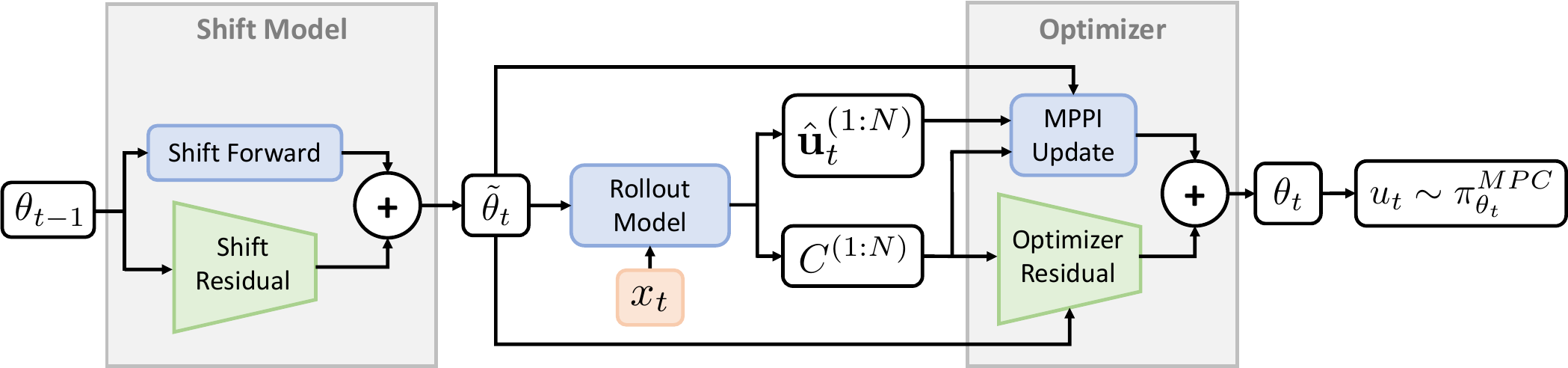}
\caption{The \dmpo architecture consists of two learnable modules, the shift model and optimizer. The fixed rollout model module performs rollouts of the sampled control sequences.}
\label{fig:arch}
\vspace{-3ex}
\end{figure*}

\subsection{Learned Optimization for MPC}
Given these approximations, the performance of MPC algorithms depends heavily on the computational budget available.
And even with sufficient resources, MPC can be sub-optimal relative to a globally optimal policy due to how we optimize our policy at each time step.
The update rule in \Cref{eq:mppi_mean_update} and \Cref{eq:mppi_variance_update} does not optimality make use of the information from the sampled trajectories.
One reason is that the update at each time step and for each control input is independent and are only coupled together via the weights from \Cref{eq:mppi_weights}.
Moreover, taking a weighted sum of the samples is a fairly simple operation and may potentially throw away useful information.
The choice of a static step size may also be limiting, and how much we update our policy may be better treated in an adaptive way.

In general, there is structure in the problem which we can leverage to construct a better update rule.
Rather than hand-design the optimization algorithm, we can \emph{learn} it via experience by parameterizing the update rule.
In the context of MPC, Sacks et al. \cite{sacks2022learning} proposed an update of the form
\begin{equation}
\thetabold_t = m_{\phi}(\tilde \thetabold_t, C_t^{(1:N)}),
\label{eq:l2o-mpc}
\end{equation}
where $C_t^{(i)} = C(\hat \xbold_t^{(i)}, \hat \ubold_t^{(i)})$ and $\phi$ are the network parameters.
Unlike the \mppi update, \Cref{eq:l2o-mpc} jointly optimizes all parameters across controls and time step.
For this to work, they generate samples from a standard Gaussian and keep them fixed.
They then use the reparameterization trick to shift and scale these samples by the current mean and standard deviation, respectively.
This makes all samples a deterministic function of our current policy parameters, which is already provided to the network.
This means we can remove the samples from the update altogether.

\section{Deep Model Predictive Optimization}
\subsection{Problem Formulation}
In \dmpo, we learn an update rule of the form in \Cref{eq:l2o-mpc} with RL, treating MPC as a structured policy class.
However, the common shift-forward operation used to warm-start the optimization only works well when problems at adjacent time steps are similar, which may not be true if there are substantial perturbations.
This choice of warm-start is also independent of the value of each decision variable.
Therefore, \dmpo additionally learns a shift model of the form
\begin{equation}
\tilde \thetabold_{t+1} = \Phi_{\phi}(\thetabold_t).
\end{equation}
Such a shift model is joint across time steps and control dimensions.
It also provides us with a parameter-dependent shift operation, which can take advantage of structure and the types of perturbations known to effect the system.

To learn these components with RL, we need to actually consider two different policies.
There is the local policy output by MPC at each time step, $\uhatbold_t \sim \pibold_{\thetabold_t}^{MPC}(\cdot)$, which we call the optimizee.
Additionally, the optimizer defines a policy over the parameters of the optimizee, $\thetabold_t \sim \pibold_\phi(\tilde\thetabold_t, C_t^{(1:N)})$.
From the perspective of RL, $\pibold_\phi$ is the policy we wish to learn, while $\pibold_{\thetabold_t}^{MPC}$ is part of the environment dynamics.
This means we have to consider both the state of the system and optimizer, forming an auxiliary MDP $\hat \Mcal = (\Hcal, \Theta, \hat P, r, \hat \rho_0, \gamma)$.
We have auxiliary states $h \in \Hcal$, where $h_{t} = (x_t, \thetabold_{t-1})$, actions $\thetabold_t \in \Theta$, which are the optimizer outputs, and dynamics
\begin{equation}
h_{t+1} = 
\begin{bmatrix}
x_{t+1} \\ \thetabold_{t}
\end{bmatrix} 
\sim \hat P(\cdot | h_t, \thetabold_{t})
=
\begin{bmatrix}
P(\cdot | x_t, u_t) \\
\delta(\thetabold_t)
\end{bmatrix},
\end{equation}
where the control of the original system, $u_t$, is a function of the auxiliary action, $\thetabold_{t}$, via 
\begin{equation}
u_t = \uhat_t,\quad 
\uhatbold_t \sim \pibold_{\thetabold_t}(\cdot).
\label{eq:optimizee_policy}
\end{equation}
From the perspective of the original MDP, the current parameters of the MPC policy actually form a recurrent state.
Therefore, even if \dmpo is parameterized with feedforward networks, we are still effectively forming a recurrent policy.
The initial state distribution $\hat \rho_0$ now samples both an initial system state and set of optimizer parameters.
And the reward function is still defined on the original state-control space, $r(x_t, u_t)$, but now $u_t$ is a function of the optimizer action $\thetabold_t$ due to \Cref{eq:optimizee_policy}.
Finally, in actor-critic algorithms, we also simultaneously learn the value function of the policy, which in this case is the optimizer, $\pibold_\phi$.
This means the critic should be defined on the auxiliary state, $V^{\pi}(h_t)$, making it a function of both the state of the system and optimizer.
Note that the optimizer never directly receives the system state $x_t$. 
While we could condition on state as well, we found that this actually hurts generalization performance, especially for sim-to-real transfer.
Intuitively, there may be multiple states for which we have the same cost distribution.
This means that the optimizer may encounter out-of-distribution states during testing which have in-distribution trajectory costs.

\subsection{Algorithm Overview}
A valid instantiation of \dmpo is to learn both the optimizer and shift model from scratch.
Instead, we learn residuals on the \mppi update and shift-forward operation.
While the closed-form \mppi update may be sub-optimal, it still can be fairly robust and generalize well to different tasks.
By learning each component as a residual operation, it can alter the reward landscape in a way that can simplify exploration.
If the \mppi controller is already good, it allows us to potentially inherit its robustness and generalization capabilities, with the residual providing small corrections.
But even if the proposed \mppi update is far from optimal, such as in the case when the controller has access to few samples, it can still provide a hint about a good direction to search.
We illustrate each module of \dmpo in \Cref{fig:arch}, which we describe below:

\textbf{Shift Model.}\ 
Let us define $\tilde \thetabold_t = (\tilde \mubold_t, \tilde \Sigmabold_t)$ as the shifted parameters of our optimizee policy.
Then we implement the shift model, $\Phi_{\phi}$, as a residual update:
\begin{equation}
\begin{gathered}
\hat \mubold_t, \hat \Sigmabold_t = \Phi_{\phi}(\thetabold_t)\\
\tilde \mubold_t = \mubold_t^{SHIFT} + \hat \mubold_t,\quad
\tilde \Sigmabold_t = \Sigmabold_t^{SHIFT} \odot \hat \Sigmabold_t,
\end{gathered}
\end{equation}
where $\mubold_t^{SHIFT}$ and  $\Sigmabold_t^{SHIFT}$ are the mean and covariance following the normal shift forward operation and $\odot$ is the Hadamard product.
While both updates could be additive, we found that the multiplicative update for the covariance worked better in practice.

\textbf{Rollout Model.}\ 
We use fixed samples from a standard Gaussian, scaling and shifting them by $\tilde \Sigmabold_t$ and $\tilde \mubold_t$, respectively, to get $\uhatbold_t^{(1:N)}$.
Additionally, we always include the current mean as one of the samples.
After this reparameterization, we roll out the open-loop control sequences  with our dynamics model to get a set of $N$ scalar total trajectory costs concatenated into a vector $C^{(1:N)}$.
Since we also run the \mppi update, we compute weights using \Cref{eq:mppi_weights}.
Importantly, this is the only module which makes use of the current system state $x_t$.

\textbf{Optimizer.}\ 
We compute the normal \mppi update using \Cref{eq:mppi_mean_update} and \Cref{eq:mppi_variance_update} to get $\thetabold_t^{MPC} = (\mubold_t^{MPC}, \Sigmabold_t^{MPC})$.
The \dmpo update for the mean is then
\begin{equation}
\begin{gathered}
\hat \mubold_t, \gbold_t, \sigmabold_t^\mu = m^{\mu}_{\phi}(\tilde \thetabold_t, C_t^{1:N})\\
\mubold_t = (1-\gbold_t) \odot \mubold_t^{MPC} + \gbold_t \odot \hat \mubold_t,
\end{gathered}
\end{equation}
where $\gbold_t$ is a gating term, which has the same dimension as the mean and is passed through a sigmoid to ensure it is between zero and one.
It allows the network to modulate how much it relies on MPC versus the network output.
Since we are using PPO \cite{schulman2017proximal} to train the components of \dmpo, we actually need a distribution over proposed mean and covariance updates.
Therefore, the network also outputs a standard deviation $\sigmabold_t^\mu$, which then defines our optimizer policy for the mean, $\pibold_{\phi}^{\mu} = \Ncal(\mubold_t, \sigmabold_t^\mu)$.
During training, we sample the mean update from this policy, but simply use the mean at test time.
Similarly, for the covariance matrix:
\begin{equation}
\begin{gathered}
\hat \Sigmabold_t, \sigmabold_t^\Sigma = m^{\Sigma}_{\phi}(\tilde \thetabold_t, C_t^{1:N}),\quad 
\Sigmabold_t = \Sigmabold_t^{MPC} \odot \hat \Sigmabold_t,
\end{gathered}
\end{equation}
where the optimizer scales the covariance matrix proposed by MPC.
Our optimizer policy for the covariance is then $\pibold_{\phi}^{\Sigma} = \Ncal(\Sigmabold_t, \sigmabold_t^\Sigma)$.
It is also important to note that these mean and covariance updates jointly depend on all the current parameters values, $\tilde \thetabold_t$.
This means that we consider the current covariance while updating the mean, and vice versa.

\section{Related Work}

\textbf{Learned Optimization.}\
There is a large body of work on L2O in the context of training neural networks \cite{andrychowicz2016learning, ravi2016optimization, lv2017learning, chen2020training, flennerhag2021bootstrapped, yang2023learning, chen2022scalable, metz2021training, metz2020tasks, wichrowska2017learned, metz2019understanding, li2016learning, li2017learning}.
Another thrust learns how to perform reinforcement learning more efficiently \cite{lu2022discovered, oh2020discovering, kirsch2019improving, oh2020discovering, houthooft2018evolved, co2021evolving, wang2016learning}.
These approaches still ultimately use standard stochastic gradient descent optimizers to update the policy and value functions.

\textbf{Combining MPC with Learning.}\ 
Common strategies to boosting the performance of MPC involve learning a dynamics model \cite{erickson2018deep, williams2017information, lenz2015deepmpc, kocijan2004gaussian, fu2016one, chua2018deep, nagabandi2018neural, finn2017deep, wahlstrom2015pixels, watter2015embed, banijamali2018robust, ebert2018visual, ha2019adaptive, hafner2019learning}, terminal value functions \cite{zhong2013value, rosolia2017learning, lowrey2018plan, bhardwaj2020blending, bhardwaj2020information}, cost-shaping terms \cite{tamar2017learning}, the entire controller end-to-end \cite{amos2020differentiable, karkus2017qmdp, okada2017path, amos2018differentiable, okada2018acceleration, pereira2018mpc, tamar2017learning}, or improving the sampling distribution \cite{power2021variational, power2022variational, sacks2023learning, okada2020variational, lambert2020stein, power2023constrained, asmar2023model, yin2022trajectory, mohamed2023towards, honda2023stein}.
However, these methods all leverage the structure of the optimization solver, learning components of the model or objective, rather than training a new update rule.
Sacks et al. \cite{sacks2022learning} explored learned optimization in the context of MPC.
However, they do not learn a shift model or a residual, and their optimizer is trained with imitation learning.
This limits their performance to the quality of the expert, which was an MPC controller with many samples.

\textbf{Residual Policy Learning.}\ 
Prior work has explored learning residual policies on a hand-designed controller with RL  \cite{lee2021bayesian, johannink2019residual, silver2018residual, yang2023continuous,yang2023cajun}.
However, unlike \dmpo, these methods do not leverage structure in the policy class.
Additionally, there is nothing specific in \dmpo necessitating a residual update. 

\textbf{MPC and RL for Quadrotor Control.}\ 
Researchers have successfully deployed sampling-based \cite{pravitra2020L1, lee2020aggressive, huang2023datt} and gradient-based \cite{hanover2021performance, sun2022comparative, kaufmann2020deep, zhang2019monocular} MPC on quadrotors.
A growing body of literature has applied RL for quadrotor stabilization \cite{molchanov2019sim, zhang2023learning, hwangbo2017control}, trajectory tracking \cite{kaufmann2022benchmark, huang2023datt}, and high-speed drone racing \cite{kaufmann2023champion}.
More closely related to our work, Romero et al. \cite{romero2023actor} embed differentiable MPC \cite{amos2018differentiable} into an actor-critic pipeline.
Song et al. \cite{song2022policy} use RL to tune the hyperparameters of MPC in an adaptive, state-dependent fashion.
However, both treat the MPC optimizer as a fixed component, instead learning how to tweak its hyperparameters and objective.

\begin{figure*}[t]
\vspace{2ex}
\centering
Unperturbed \hspace{20ex}
Wind + Drag Plate \hspace{22ex}
Setup\\
    \hspace{-2ex}
    \begin{subfigure}[t]{0.31\textwidth}
        \centering
        \includegraphics[width=1\textwidth]{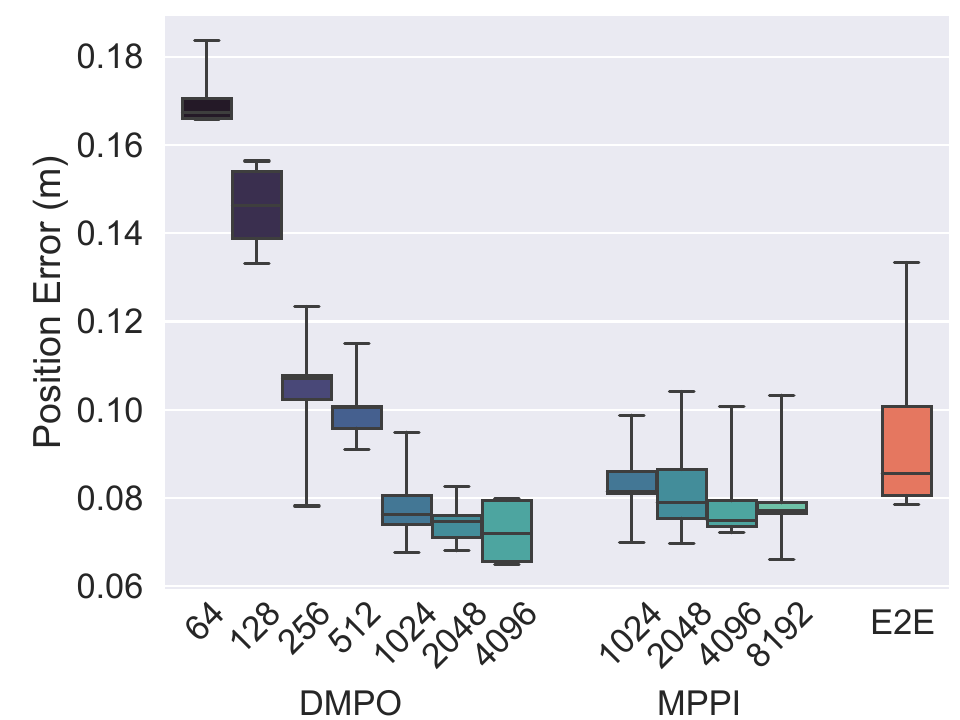}
    \end{subfigure}
    \begin{subfigure}[t]{0.31\textwidth}
        \centering
        \includegraphics[width=1\textwidth]{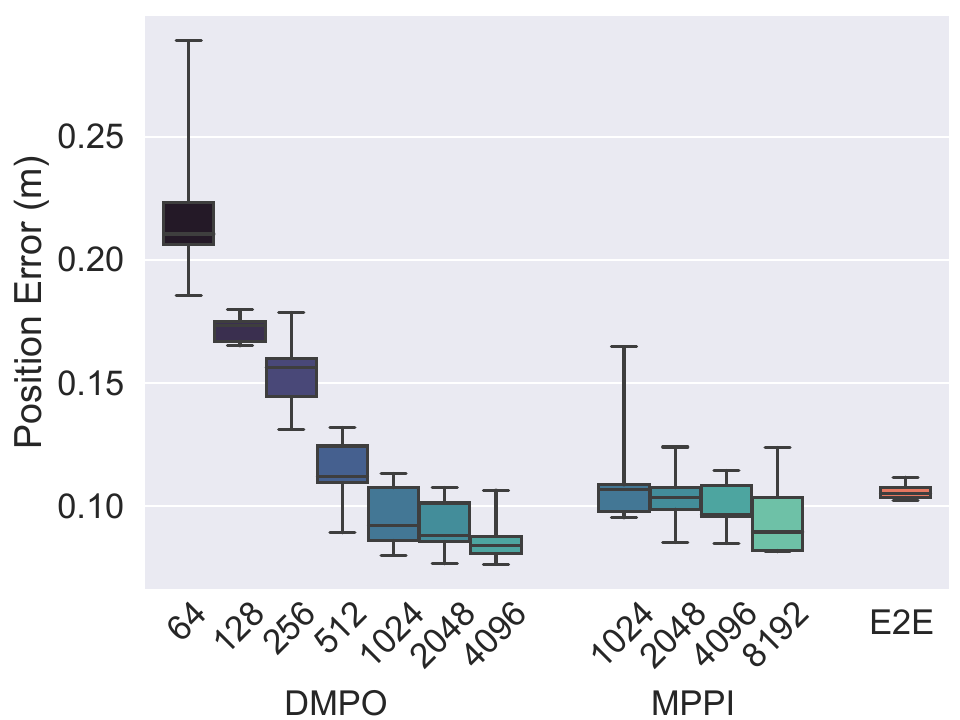}
    \end{subfigure}
    \hspace{2ex}
    \begin{subfigure}[t]{0.275\textwidth}
        \centering
        \raisebox{4.7ex}{
        \includegraphics[width=1\textwidth]{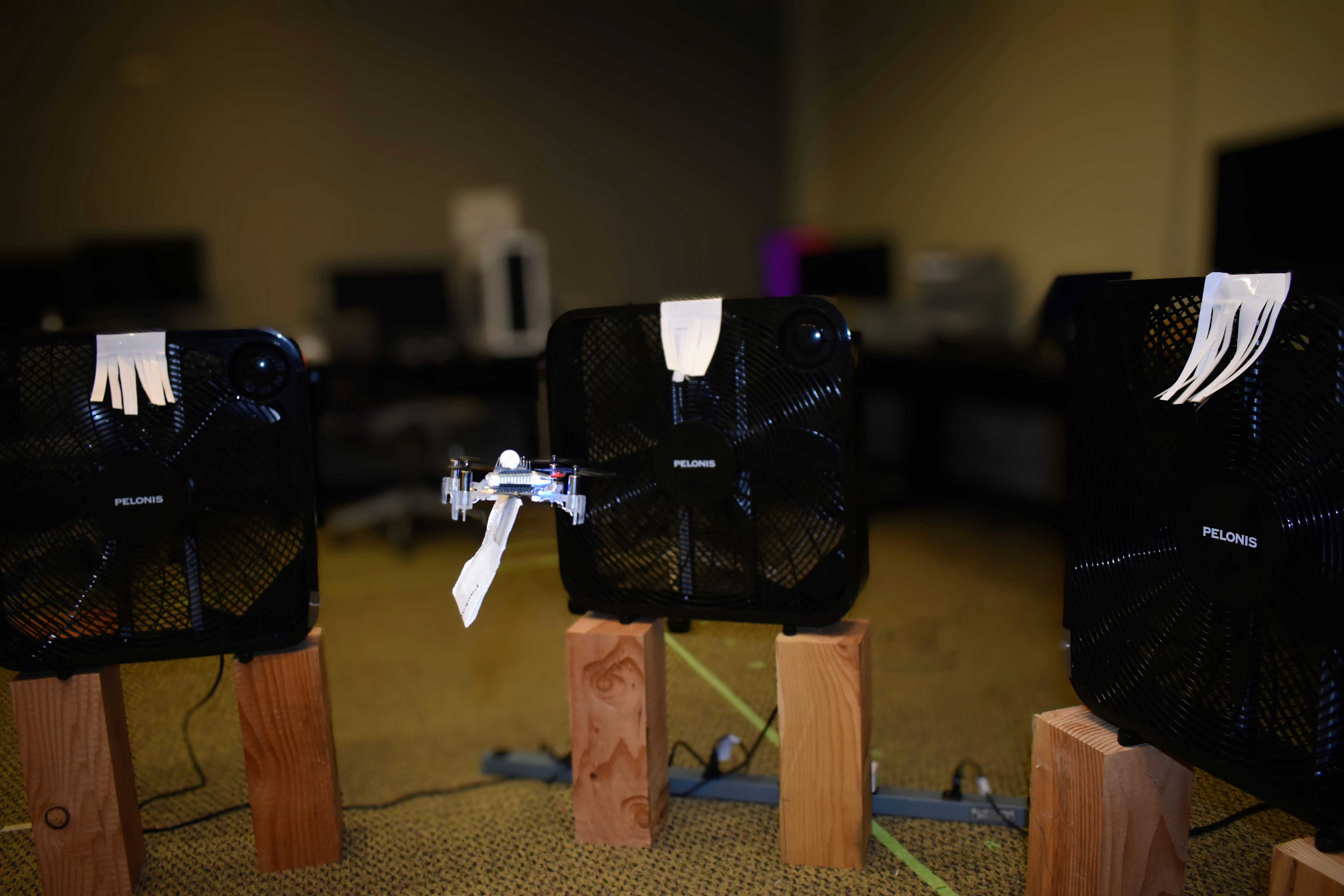}
        }
    \end{subfigure}
    \caption{Position tracking error of \dmpo versus \mppi and \ete on tracking random infeasible zig-zag trajectories without any environmental disturbances (left) and with an attached plate and wind (middle), with the setup shown on the right.}
    \label{fig:zz_costs}
    \vspace{-3ex}
\end{figure*}

\begin{figure*}[t]
    \centering
    \includegraphics[width=1\textwidth]{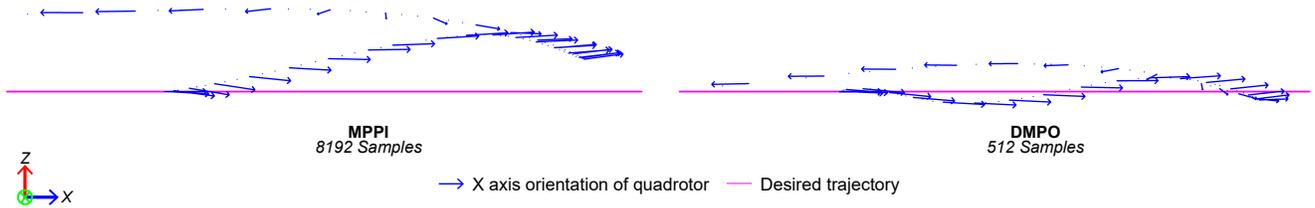}
    \caption{Example zig-zag trajectory with a $180^o$ yaw flip performed by \mppi (8192 samples) and \dmpo (512 samples).}
    \label{fig:zzyaw}
    \vspace{-3ex}
\end{figure*}

\section{Experiments}
\subsection{Task and Implementation Details.}
\textbf{Quadrotor Trajectory Tracking.}\
We perform all evaluations on a quadrotor trajectory tracking problem in which the desired trajectories are infeasible zig-zags with and without yaw flips.
These zig-zags linearly connect a series of random waypoints, while the yaw flips are a $180^o$ change to the desired yaw at each waypoint.
For the real hardware experiments, we use the Bitcraze Crazyflie 2.1 equipped with the longer 20 mm motors from the thrust upgrade bundle.
State estimation for position and velocity is provided by an OptiTrack motion capture system, while the Crazyflie provides orientation estimates via a 2.4 GHz radio.
An offboard computer receives the state estimates and runs all controllers at a rate of 50 Hz.
All controllers operate on desired body thrust $f_{des}$ and angular velocity $\omegabold_{des}$ commands.
We convert these commands to motor thrusts using a low-level controller.

For training and the MPC dynamics model, we use the following dynamics with a $\text{dt} = 0.02$:
\begin{equation}
\begin{aligned}
\dot \pbold = \vbold,\quad 
m\dot \vbold = m\gbold + \Rbold \ebold_3 f,\quad
\dot \Rbold = \Rbold S(\omega),
\end{aligned}
\end{equation}
where $\pbold, \vbold, \gbold \in \real^3$ are the position, velocity, and gravity vectors in the world frame, $\Rbold \in \text{SO}(3)$ is the attitude rotation matrix, $\omegabold \in \real^3$ is the angular velocity in the body frame, $S(\cdot): \real^3 \rightarrow \text{so}(3)$ maps a vector to its skew-symmetric matrix form, $\ebold_3$ is a unit vector in the Z direction, and $m$ is the mass.
The state of the system is then $\xbold = (\pbold, \vbold, \qbold, \omegabold)$, where $\qbold$ is the quaternion representation of $\Rbold$, and $\ubold = (f_{des}, \omegabold_{des})$.
Rather than explicitly model the angular velocity dynamics and low-level controller, we convert $\ubold$ to the actual thrust $f$ and angular velocity $\omegabold$ using a first-order time delay model:
\begin{equation}
\begin{aligned}
\omegabold_t &= \omegabold_{t-1} + k(\omegabold_{des} - \omegabold_{t-1}) \\
f_t &= f_{t-1} + k(f_{des} - f_{t-1}).
\end{aligned}
\end{equation}
The cost function includes terms defined on position and orientation tracking performance.
Additionally, a control penalty was necessary for sim-to-real transfer.
Without it, \dmpo would exploit the simulator and learn aggressive commands which are difficult to perform on the real system.

\textbf{Hyperparameters and Training.}\ 
The optimizer, shift model, and value function were all parameterized with multi-layer perceptrons (MLPs) with a single hidden layer of 256 and \relu\ activation functions implemented in PyTorch \cite{paszke2019pytorch}.
Since the value function operates on the full auxiliary state, we needed to make it aware of the reference trajectory.
Therefore, we give it the desired trajectory for the next 32 time steps with a stride of 4, forming a 56-dimensional conditioning vector.
However, we note that this is not necessary for the optimizer or shift model, as they do not operate on states.
For the network initialization, we set the last layer of each MLP to have a weight distribution of $\Ncal(0, 0.001)$.
This made each residual term effectively zero, allowing us to start with the \mppi\ update and shift-forward operation for warm-starting.

We trained the optimizers with PPO \cite{schulman2017proximal} and Generalized Advantage Estimation (GAE) \cite{schulman2015high} on an NVIDIA RTX 3080 GPU with a $\gamma = 0.99$ and $\lambda = 0.95$.
To update \dmpo, we used Adam \cite{kingma2014adam} with a learning rate of $10^{-6}$ and $10^{-4}$ for the actor and critic, respectively.
Learning the \dmpo residual optimizers only took up to 1000 iterations of PPO to achieve good performance.
To improve performance, we used domain randomization (DR) \cite{tobin2017domain, peng2018sim, chen2021understanding} on the mass, randomly scaling it by a factor in $[0.7\times, 1.3\times]$, and the delay coefficient $k$, selecting it in $[0.2, 0.6]$.
We also applied a constant force perturbation with a randomized direction and magnitude at the beginning of each episode in $[-3.5\ N, 3.5\ N]$.

\textbf{Baselines.}\ 
Our two baselines are \mppi\ and an end-to-end (\ete) 3-layer MLP policy operating on states,  conditioned on desired trajectories. The desired trajectories consist of the 10 desired positions up to $0.6$ seconds in the future, evenly spaced in time.
\ete was also trained with PPO using DR, but with a learning rate of $3\times 10^{-4}$ and took about $10^7$ iterations to converge.
We used a custom implementation of \mppi\ in PyTorch \cite{paszke2019pytorch}, which used Halton sequences \cite{halton1964algorithm} for generating the fixed samples from a standard Gaussian.
We tune all hyperparameters of the \mppi controller using a grid search in simulation on a fixed set of desired trajectories.

\begin{figure*}[t]
\vspace{2ex}
\centering
\begin{minipage}[t]{\textwidth}
    \hspace{2ex}
    \raisebox{12ex}{\rotatebox[origin=t]{90}{Unperturbed}}
    \begin{subfigure}[t]{0.30\textwidth}
        \centering
        \includegraphics[width=1\textwidth]{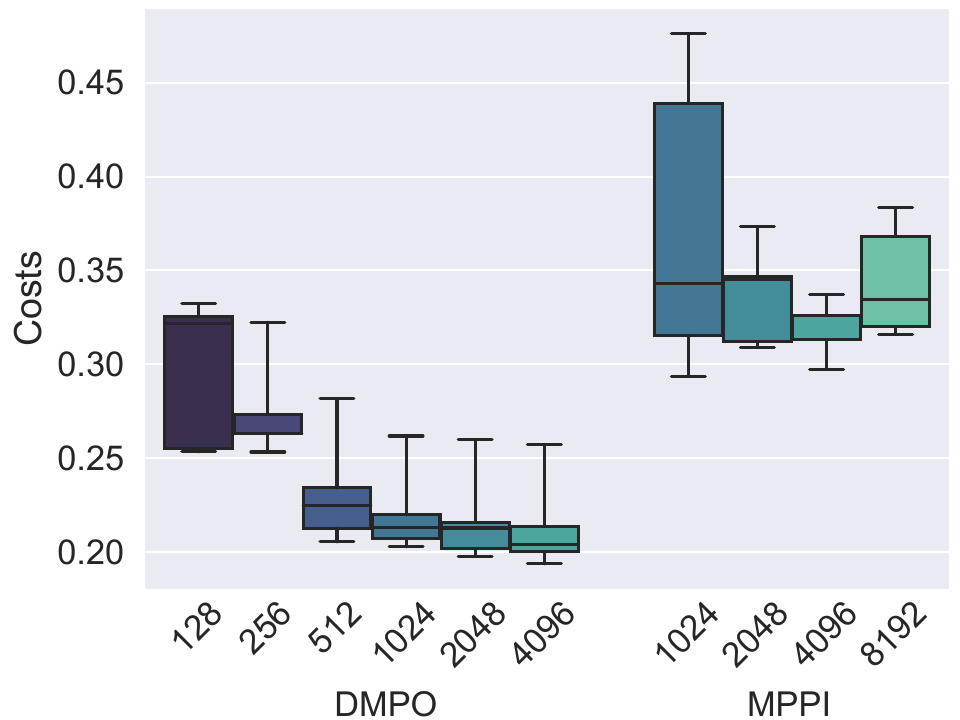}
    \end{subfigure}
    \begin{subfigure}[t]{0.30\textwidth}
        \centering
        \includegraphics[width=1\textwidth]{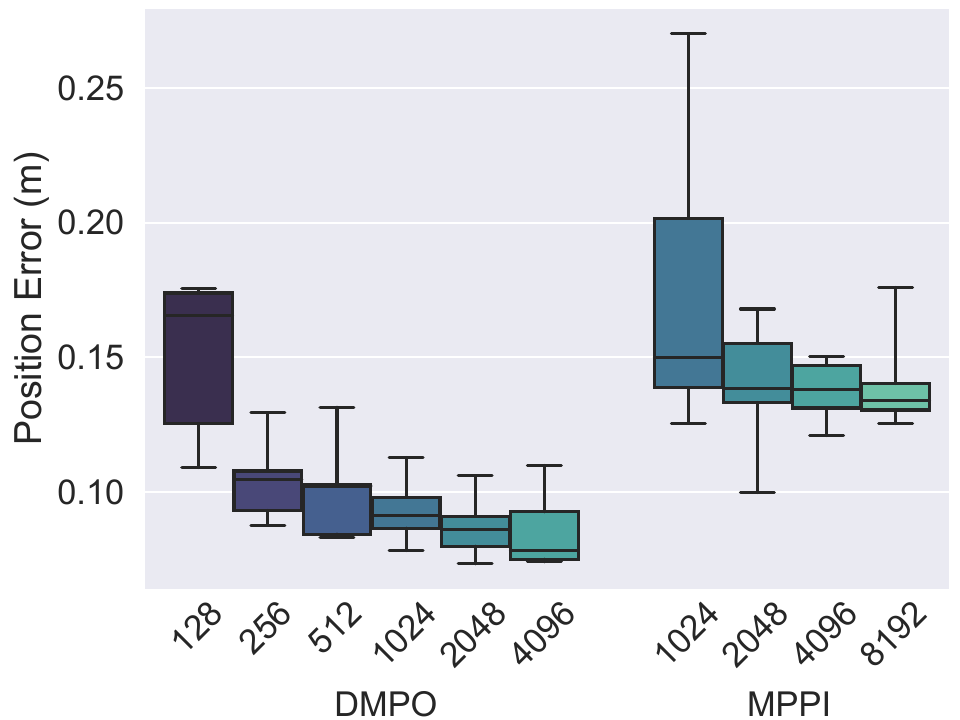}
    \end{subfigure}
    \begin{subfigure}[t]{0.30\textwidth}
        \centering
        \includegraphics[width=1\textwidth]{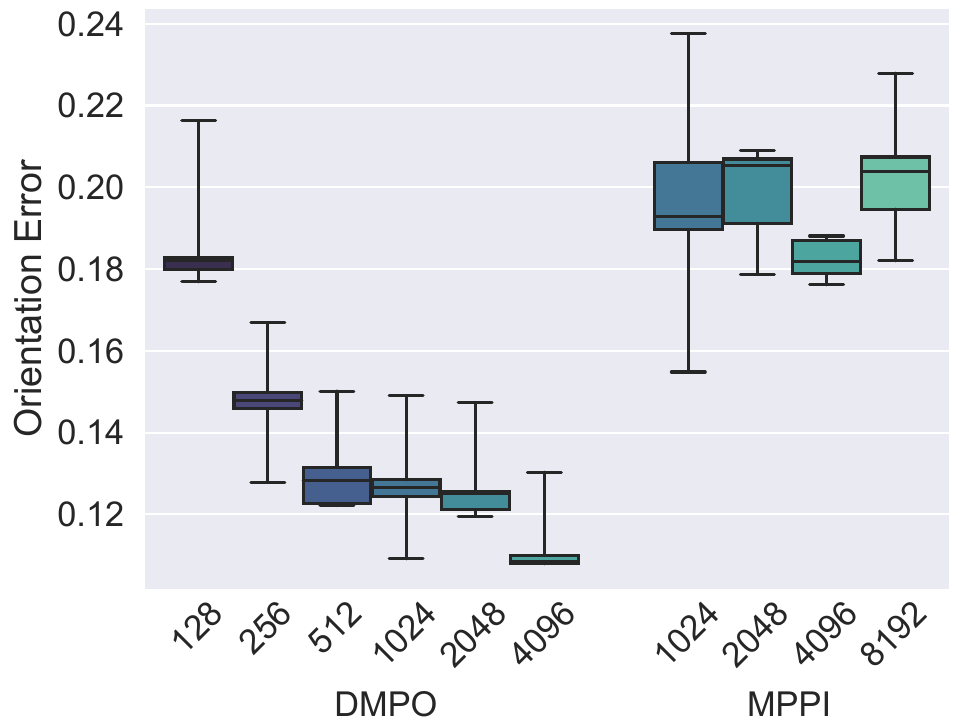}
    \end{subfigure}
    \vspace{2ex}
\end{minipage}
\begin{minipage}[t]{\textwidth}
\hspace{2ex}
    \raisebox{12ex}{\rotatebox[origin=t]{90}{Wind + Drag Plate}}
    \begin{subfigure}[t]{0.30\textwidth}
        \centering
        \includegraphics[width=1\textwidth]{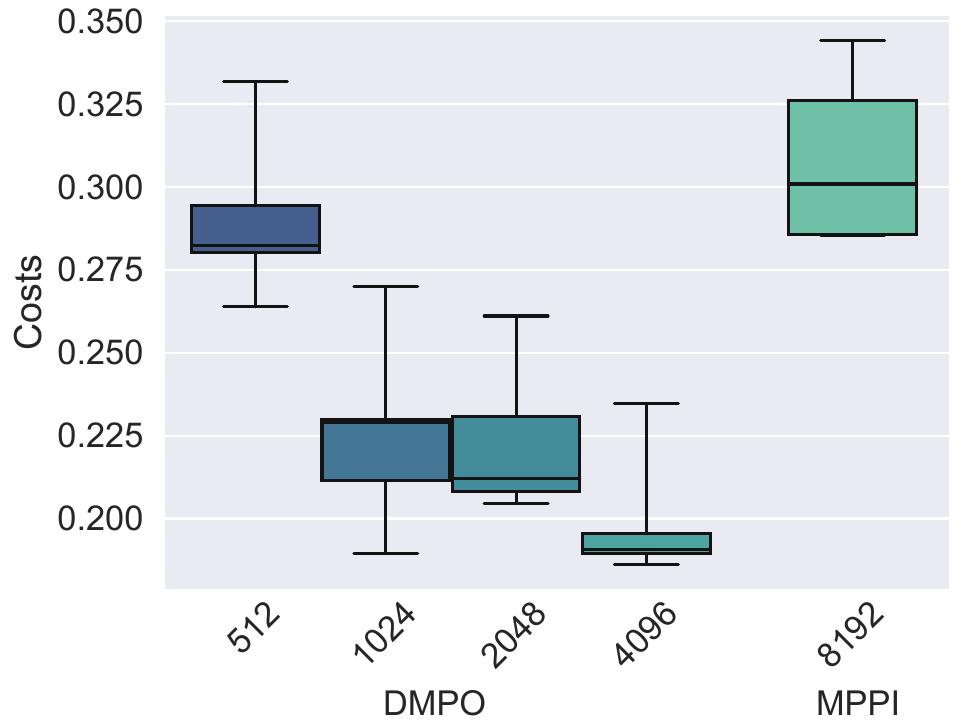}
    \end{subfigure}
    \begin{subfigure}[t]{0.30\textwidth}
        \centering
        \includegraphics[width=1\textwidth]{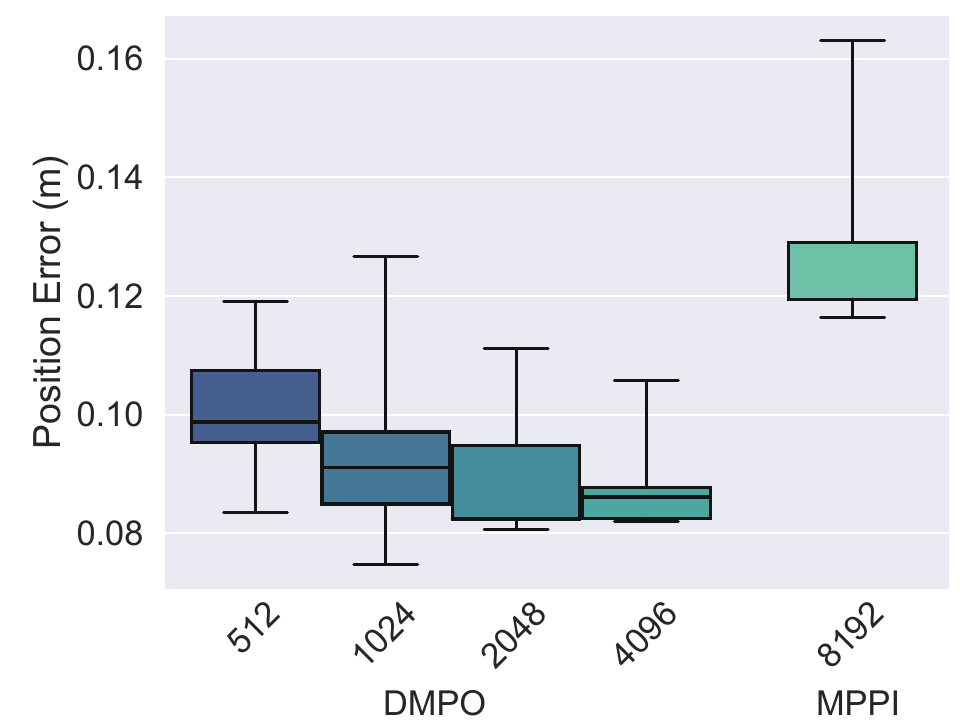}
    \end{subfigure}
    \begin{subfigure}[t]{0.30\textwidth}
        \centering
        \includegraphics[width=1\textwidth]{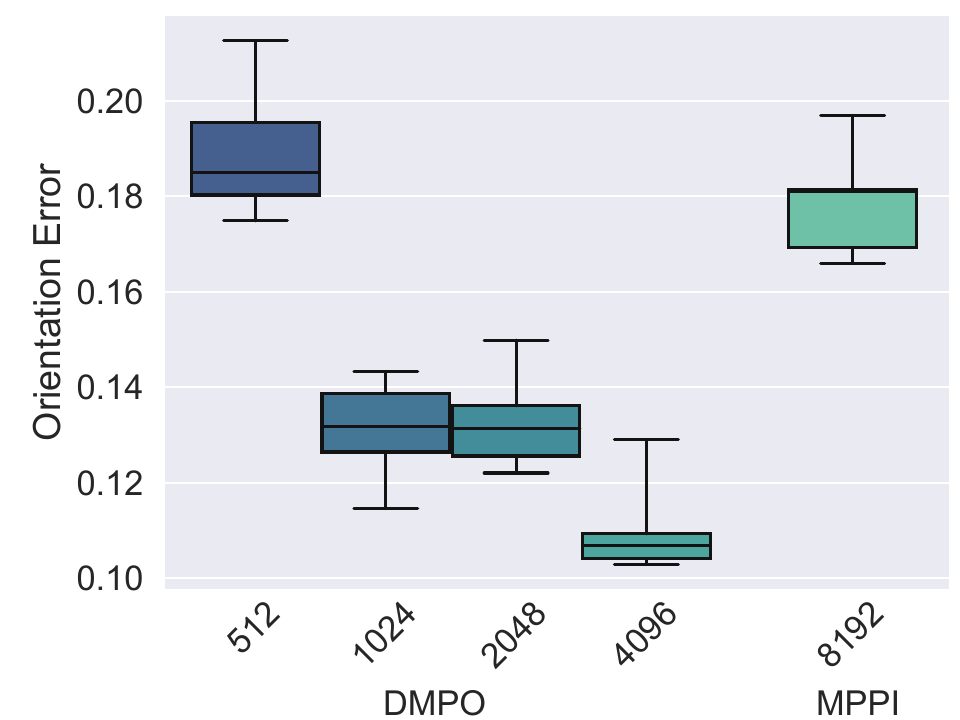}
    \end{subfigure}
\end{minipage}
\caption{Total cost (left), position error in meters (middle), and orientation error (right) of \dmpo versus \mppi on tracking random yaw flip trajectories without any environmental disturbances (top) and with an attached plate and wind (bottom).}
\label{fig:costs}
\vspace{-3ex}
\end{figure*}


\subsection{Tracking Performance on Infeasible Zig-Zags}
We begin by evaluating the performance of \dmpo compared to \mppi and \ete on random zig-zag trajectories without the presence of additional disturbances.
For each controller, we evaluate the performance across 5 different fixed trajectory seeds.
\Cref{fig:zz_costs} (left) reports the position tracking error of each controller, varying the number of samples for \dmpo and \mppi.
The box plots represent the median and quartiles of the total episode costs.
Given 512 or less samples, \mppi would consistently crash, while \dmpo is able to successfully complete the task with as few as 64 samples.
We found that increasing the number of samples for \mppi only helps to a point, after which performance can suffer.
However, increasing the number of samples generally improves the median performance of \dmpo.
And \dmpo with 1024 samples outperforms the best \mppi controller (with 4096 samples) by 7\% and \ete by 19\% in terms of the median error.

In order to gauge whether \dmpo retains the robustness of MPC, we test the Crazyflie in a scenario with an unknown wind field generated by three fans.
Additionally, we attached a soft cardboard plate hanging below the chassis, which creates drag and adds additional mass (see the right of \Cref{fig:zz_costs}).
The combination of the fans and cardboard plate creates highly dynamic and state-dependent disturbances which are not encountered during training.
We report the results of these perturbations on the right of \Cref{fig:zz_costs}.
The performance of all three controllers got worse, although \dmpo can still remain in the air with as few as 64 samples.
\dmpo with 1024 samples nearly matches the performance of the best \mppi controller with 8192 samples.
And it still surpasses the performance of \ete by 14\%, with the improvement growing with samples.

\subsection{Tracking Performance on Yaw Flips}
Next, we evaluate performance on zig-zags with yaw flips.
The \ete baseline could not successfully transfer to the real system on this task.
The top row of \Cref{fig:costs} reports the performance of \dmpo and \mppi without the presence of additional disturbances.
Again, with 512 samples or fewer, \mppi would consistently crash, while \dmpo with as few as 128 can successfully stay in the air.
At 256 samples, \dmpo outperforms the best \mppi controller with 4096 samples by over 27\%.
And in this much harder scenario, \dmpo with 4096 samples is 64\% better than \mppi.
Breaking down the cost, we see that \dmpo with 256 or more samples does a much better job in tracking both desired position and orientation.
\Cref{fig:zzyaw} illustrates an example trajectory and how \dmpo has much better position tracking (especially in the Z direction) and rotates more rapidly to improving orientation tracking.

\begin{table}[t]
\normalsize
\centering
\caption{Relative memory usage between \mppi with 4096 samples and \dmpo.}
\begin{tabular}{c|c|c|c|c}
\hrulethick
\textbf{\# \dmpo Samples} & 256 & 512 & 1024 & 2048\\
\hline
\textbf{Memory Reduction} & $4.3\times$ & $3.1\times$ & $1.9\times$ & $1.1\times$\\
\hrulethick
\end{tabular}
\label{tab:memory}
\vspace{-3ex}
\end{table}

We report the perturbation results in the bottom row of \Cref{fig:costs}.
In this scenario, \mppi needed 8192 samples to avoid crashing, while \dmpo remained robust at 512 samples, outperforming \mppi by 7\%.
And given 4096 samples, \dmpo is over 57\% better than \mppi.
We again see that position error is substantially lower for \dmpo.
In contrast, \dmpo at 512 samples was slightly worse than \mppi for orientation error.
Yet, \dmpo quickly got better at tracking orientation with 1024 or more samples.
Comparing \dmpo in this case to the unperturbed scenario, we see that its performance is markedly similar.
In fact, it only got worse by about 7\% on average, while \mppi incurred about 11\% more cost.
The median position and orientation errors are only slightly larger with disturbances, except for the 512-sample \dmpo. 
Together, these results indicate the zero-shot generalization capability of \dmpo in the presence of unknown disturbances.
Additional results  can be found at \href{https://tinyurl.com/mr2ywmnw}{https://tinyurl.com/mr2ywmnw}.

\subsection{Compute Requirements}
\dmpo with 256 samples is $1.2 \times$ faster than \mppi with 4096 samples on our offboard computer while outperforming it on the yaw flip task.
And the savings may be even greater for on-board compute which is more constrained.
We report the memory usage of \dmpo for various samples compared to \mppi with 4096 samples in \Cref{tab:memory}.
With 256 samples, \dmpo requires $4.3\times$ less memory.
Altogether, this means that we can achieve better performance while using less compute and memory compared to \mppi.

\section{Conclusion}
We devised \dmpo, a method for jointly learning the optimizer and warm-starting procedure for MPC.
By framing the optimizer as a policy in an auxiliary MDP, we showed how MPC could be treated as a structured policy class and learned via MFRL.
We evaluated \dmpo on a real quadrotor platform tracking infeasible zig-zag trajectories and showed it can outperform \ete and \mppi controllers with far fewer samples.
And \dmpo is even more robust than \mppi to unseen disturbances, such as unknown wind fields and an attached cardboard drag plate.
Moreover, since \dmpo can accomplish this level of performance with fewer samples, it can save up to $4.3 \times$ memory and reduce runtime by $1.2\times$ compared to \mppi.
This indicates \dmpo is a viable strategy to leverage the robustness of MPC while improving upon these hand-designed controllers and better match the optimal policy.


\bibliography{references}
\bibliographystyle{IEEEtran}

\end{document}